\documentclass[lettersize,journal]{IEEEtran}
\usepackage{amsmath,amsfonts}
\usepackage{algorithmic}
\usepackage{algorithm}
\usepackage{array}
\usepackage[caption=false,font=normalsize,labelfont=sf,textfont=sf]{subfig}
\usepackage{textcomp}
\usepackage{stfloats}
\usepackage{url}
\usepackage{verbatim}
\usepackage{graphicx}
\usepackage{cite}
\usepackage{makecell}
\usepackage{graphicx}
\usepackage{textcomp}
\usepackage{array} 
\usepackage{longtable}
\usepackage{booktabs}
\usepackage{float}
\usepackage[pagebackref,breaklinks,colorlinks,citecolor=blue]{hyperref}
\begin{document}
	\title{Motion State: A New Benchmark Multiple Object Tracking}
	
	\author{
		Liao Pan,
		Yang Feng,
		Wu Di,
		Liu Bo,
		Zhang Xingle
		% <-this % stops a space
		\thanks{  Liao Pan, Yang Feng, Wu Di, Liu Bo, and Zhang Xingle are with Key Laboratory of Information Fusion, School of Automation, Northwestern Polytechnical University,Xi'an,China.	(e-mile: liaopan@mail.nwpu.edu.cn,yangfeng@nwpu.edu.cn,wu\_di821@mail.
			nwpu.edu.cn,
				503790475@qq.com, 2072458850@qq.com
			    }
			}
	
	
	
	\maketitle
	
	\begin{abstract}
			
		In the realm of video analysis, the field of multiple object tracking (MOT) assumes paramount importance, with the motion state of objects—whether static or dynamic relative to the ground—holding practical significance across diverse scenarios. However, the extant literature exhibits a notable dearth in the exploration of this aspect. Deep learning methodologies encounter challenges in accurately discerning object motion states, while conventional approaches reliant on comprehensive mathematical modeling may yield suboptimal tracking accuracy. To address these challenges, we introduce a Model-Data-Driven Motion State Judgment Object Tracking Method (MoD2T). This innovative architecture adeptly amalgamates traditional mathematical modeling with deep learning-based multi-object tracking frameworks. The integration of mathematical modeling and deep learning within MoD2T enhances the precision of object motion state determination, thereby elevating tracking accuracy. Our empirical investigations comprehensively validate the efficacy of MoD2T across varied scenarios, encompassing unmanned aerial vehicle surveillance and street-level tracking. Furthermore, to gauge the method's adeptness in discerning object motion states, we introduce the Motion State Validation F1 (MVF1) metric. This novel performance metric aims to quantitatively assess the accuracy of motion state classification, furnishing a comprehensive evaluation of MoD2T's performance. Elaborate experimental validations corroborate the rationality of MVF1. In order to holistically appraise MoD2T's performance, we meticulously annotate  several renowned datasets and subject MoD2T to stringent testing. Remarkably, under conditions characterized by minimal or moderate camera motion, the achieved MVF1 values are particularly noteworthy, with exemplars including 0.774 for the KITTI dataset, 0.521 for MOT17, and 0.827 for UAVDT.
		
	\end{abstract}
	
	
	\section{Introduction}
		The determination of whether an object is in motion, i.e., its motion state, is deemed to hold significant practical implications across various domains. For instance, in the realm of autonomous driving, moving objects often warrant greater attention than stationary ones, as foreknowledge of the motion state of objects on the road surface could potentially aid in autonomous driving decision-making. Similarly, in the field of remote sensing, the focus on moving objects supersedes that on stationary ones. Given that discerning the motion state of objects necessitates knowledge of their continuous positional coordinates, we posit that this issue constitutes a subtask within the domain of multiple object tracking (MOT).
		
		MOT stands as a pivotal cornerstone in the realm of video analysis, capturing the dedicated attention of scholars and researchers alike due to its fundamental importance in numerous real-world applications. MOT originated from the necessity to overcome the limitations inherent in traditional detection techniques, prompting initial pursuits that encompassed methodologies like image differencing, Kalman filtering, and the Hungarian algorithm, thereby laying the foundational bedrock of tracking frameworks \cite{liptonMovingTargetClassification1998}. Subsequent advancements witnessed the emergence of motion detection techniques, including the application of frame differencing, and statistical model-centric strategies, exemplified by Gaussian mixture models \cite{yiDetectionMovingObjects2013} and adaptive background modeling \cite{barnichViBeUniversalBackground2011}. However, it remains conspicuous that traditional non-deep learning methods, fixated solely on tracking moving objects, often yield suboptimal outcomes across diverse scenarios due to the intricate nature of image complexities.
		
		In the ever-evolving landscape of computer vision, a profound transformation has unfolded, primarily steered by the pervasive embrace of deep learning. This transformative wave finds its embodiment in two prominent methodologies: tracking-by-detection and joint-detection-association. Tracking-by-detection intricately harnesses the power of deep learning models for precise object detection, seamlessly integrating them with sophisticated techniques like Kalman filtering to pave the way for seamless object tracking \cite{bewleySimpleOnlineRealtime2016,aharonBoTSORTRobustAssociations2022,maggiolinoDeepOCSORTMultiPedestrian2023}. On the flip side, joint-detection-association unfolds as an end-to-end integrated tracking network \cite{Cetintas_2023_CVPR,li2023panet,wu2021track,zhang2023motrv2,wu2022improving}. The deep learning paradigm brings a tapestry of advantages to the forefront in the realm of MOT, endowing the system with unparalleled capabilities for feature learning and representation, allowing for a nuanced understanding of the semantic intricacies of tracked objects \cite{zhou2023body, guo2017efficient}.
		
		Deep learning's forte lies in its adeptness at modeling context effectively, enabling a holistic comprehension of the interplay between tracked objects and their surroundings \cite{zhou2023body}. The scalability of deep learning models in handling vast amounts of data adds another layer of strength, a crucial facet given the large-scale nature of MOT tasks. Furthermore, the principle of transfer learning accelerates the adaptation of models from one task to another, a vital asset in the ever-evolving MOT landscape.
		
		Adaptability emerges as a hallmark of deep learning, particularly in navigating the complexities of dynamic scenes marked by occlusions and intricate motion patterns. This innovative era has not only witnessed the ascendancy of tracking-by-detection but has also translated into remarkable successes across a diverse spectrum of applications \cite{sunDanceTrackMultiObjectTracking2022,milanMOT16BenchmarkMultiObject2016,dendorferMOT20BenchmarkMulti2020}.
		
		However, these neural network-based methods heavily rely on object appearance features during implementation. Consequently, they often track all objects within an image, deviating from traditional methods that specifically target moving objects. Environments such as vehicle monitoring, battlefield operations, and autonomous driving typically require a focus on moving objects. In such scenarios, acquiring information about the motion state of objects in relation to the ground poses a challenge that current methods find difficult to tackle. Furthermore, as many scenarios utilize single RGB cameras rather than stereo or RGB-D cameras, discerning object motion becomes even more challenging. Addressing this issue, this paper introduces a straightforward yet practical approach for determining object motion, adaptable across various environments.
		
		Traditional non-deep learning methods adeptly track moving targets, harnessing the physical and mathematical facets of images and relying on comprehensive mathematical models. While they might exhibit reduced accuracy in select cases, their efficacy extends across almost any environment. In contrast, deep learning-driven MOT frameworks pivot on data-driven principles. Though their performance is notable, they may falter when confronted with limited or inadequate data, leading to inaccuracies in tracking object motion estimation.

		To transcend the limitations inherent in both deep learning algorithms and traditional methods, we present a tracking approach that melds conventional mathematical modeling with data-driven deep learning methodologies. In essence, this paper unfolds with three core contributions: 
		\begin{itemize}
			\item [1)] introducing the Motion State Validation F1 Score (MVF1) as a metric to assess the model's proficiency in determining the relative motion status of objects in relation to the ground. The experimental validation of MVF1 is conducted, offering substantiating evidence for its soundness; 
			
			\item [2)] presenting a plug-and-play method for discerning the motion attributes of tracked objects; 
			
			\item [3)] proposing a model-data-driven tracking framework(MoD2T) that harmonizes traditional methods with MOT tracking frameworks, enabling the delineation of object motion states.
		\end{itemize}

	\section{RELATED WORK}
	\subsection{Motion Objects Detection Methods}
	
	%Regarding the detection of moving objects in MOT, it differs somewhat from general motion recognition, such as identifying movements like \cite{zhang2022motion}, and so on. Rather than being categorized as motion detection, it would be more appropriate to describe it as image change detection.
	
	Traditional MOT methods primarily focus on identifying moving targets, and the subsequent tracking process is similar to the SORT\cite{bewleySimpleOnlineRealtime2016}. In the early stages, image change detection methods were used for this purpose.
	
	The development of image change detection methods can be traced back several decades. The earliest methods were based on pixel value comparison to detect changes \cite{tesauro1995temporal,liptonMovingTargetClassification1998,piccardiBackgroundSubtractionTechniques2004}. These methods were simple and intuitive but sensitive to complex scenes and noise.
	
	Subsequently, methods based on pixel-level differences received extensive research attention. These methods use pixel-level statistics or thresholds to detect changes, such as pixel ratio \cite{gong2011change} and statistical parameters \cite{bouwmans2008background}. While these methods are robust to noise, they still struggle with complex textures and lighting variations \cite{panReviewVisualMoving2017}.
	
	Before the surge of deep learning, various Gaussian background model were the state-of-the-art in this field. Specifically, for a background image, the grayscale values of specific pixel points in the temporal domain follow a Gaussian distribution. In other words, for a background image $B$, the grayscale value $x$ of each point $(u,v)$ satisfies $I(x)\sim N\left( \mu ,\sigma ^2 \right)$ as shown in (1):
	
	\begin{equation}
		%	\label{deqn_ex1a}
		I(x)=\frac{1}{\sqrt{2\pi}\sigma}e^{-\frac{(x-\mu )^2}{2\sigma ^2}}
	\end{equation}
	
	In the process of extracting targets, the mean $\mu$ and variance $\sigma$ of each point in the image sequence over a period of time are calculated to create the background model. Then, for any image that contains foreground objects, each point is evaluated based on (2):
	\begin{equation}
		%	\label{deqn_ex1a}
		\frac{1}{\sqrt{2\pi}\sigma}e^{-\frac{(IG(x,y)-IB(x,y))^2}{2\sigma ^2}}>T
	\end{equation}
	If the condition is met, the point is considered a background point; otherwise, it is considered a foreground point. Here, $T$ is a constant threshold representing the gray scale value at a given point in image $IG$ and $IB$ is similar in nature.
	
	In the field of image processing, where deep learning methods dominate, motion detection methods primarily adopt sparse feature tracking\cite{zhongDetectSLAMMakingObject2018,zhang2020vdo,liu2022vslam} or dense optical flow \cite{sand2008particle,gehrig2021raft,bouwmeester2023nanoflownet}. The former is widely used in Simultaneous Localization and Mapping(SLAM), while the latter is commonly applied for overall motion estimation in images. Although each method has been proven effective for its respective applications, neither can fully model the motion in videos: optical flow cannot capture long-term motion trajectories, and sparse tracking cannot model the motion of all pixels. Recently, someone proposed a method combining the Optical flow with some other modes \cite{wang2023tracking}, and achieved good results.
	
	However, these methods still have certain limitations. Sparse feature tracking involves random selection of feature points, which may result in tracking objects with too few or no features, or unnecessary computational overhead if too many features are selected. The limitations of optical flow methods based on CNN\cite{liuEmpiricalComparisonGANs2021,li2023ecg} or generative adversarial networks (GAN)\cite{10.1007/978-3-030-58536-5_24,liu2023udf} include current challenges regarding their accuracy and universality. Traditional optical flow methods, on the other hand, exhibit high tracking accuracy in current MOT tracking frameworks.
	
	Of course, in recent years, several models based on deep learning for detecting changes in images have been proposed \cite{rahmonMotionUNetMulticue2021,shi2020change,asokan2019change}. These methods have been widely used in fields such as remote sensing and change detection. However, due to the fact that their working principle relies on static image comparison, these methods are no longer suitable for motion object detection.
	\subsection{MOT Methods And Fusion Methods}
	The developmental trajectory of MOT methodologies has been expounded upon in the preceding text and will not be extensively reiterated here. 
	
	Currently, the available MOT approaches capable of distinguishing object motion exhibit limited proliferation. A significant portion of these approaches do not solely rely on RGB image information, but rather necessitate supplementary sensing modalities such as RGB-D cameras\cite{sanchez2017robust} or various alternative sensors like LIDAR\cite{zhongDetectSLAMMakingObject2018}.
	
	The conventional MOT approach employed within this paoer shall be elucidated in the appendix. As for the deep learning-driven MOT method, this paper intends to embrace the tracking-by-detection paradigm.
	
	For the object detection stage, YOLOv8\cite{yolov8_ultralytics} will be used. YOLOv8 is a fast and accurate object detection method that divides the image into grids and predicts bounding boxes and their corresponding classes for object detection.
	
	For the subsequent tracking stage, StrongSORT\cite{duStrongSORTMakeDeepSORT2023} will be employed. StrongSORT is a deep learning-based object tracking method that combines appearance features and motion information to achieve accurate multiple object tracking.
	
	The tracking framework, YOLOv8+StrongSORT, combines the YOLOv8 object detector with the StrongSORT tracker, resulting in powerful tracking performance. YOLOv8 can rapidly and accurately detect targets in video frames and provide bounding box locations and class information for each detection. StrongSORT utilizes these detection results and combines them with the appearance features and motion information of the targets. It performs data association and trajectory prediction for target tracking and re-identification.
	
	Additionally, the paper has also experimented with other tracking methods, such as YOLOv8+ DeepOCSORT \cite{maggiolinoDeepOCSORTMultiPedestrian2023}, to explore multiple tracking approaches. Our method can be integrated with different MOT methods, and this article only selected two.
	\begin{figure*}[htbp]
		\centering
		\includegraphics[width=5in,height=2.5in]{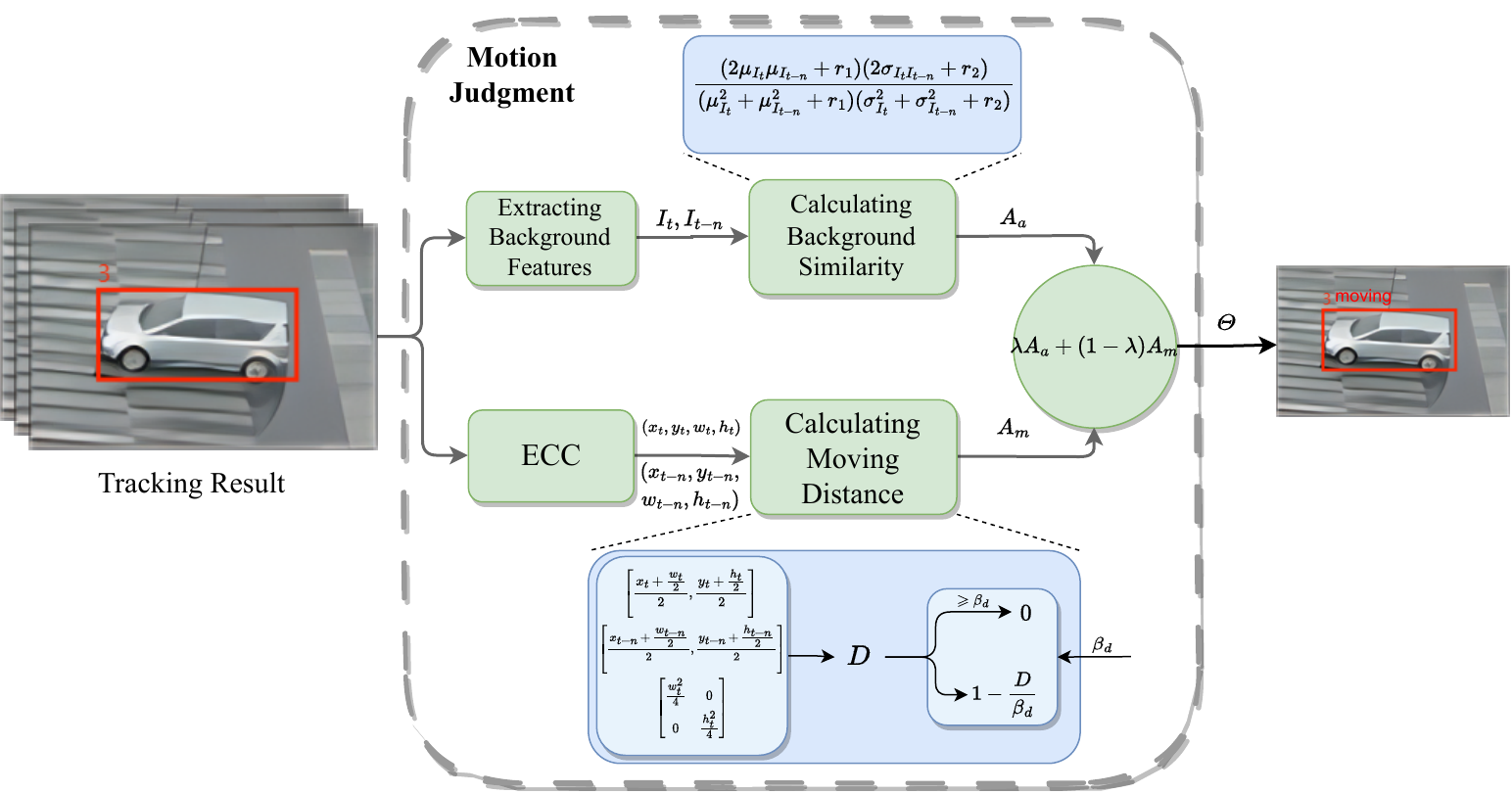}
		\caption{Flow chart of motion state judgment. The background feature extraction process is shown in Figure \ref{feature}. And, in this paper, The frame rate difference $n$ for calculating distance and background similarity is set to  3. The specific meaning of the formula and the specific meanings of each parameter can be found in section \ref{Motion Judgment}. The main purpose of this process is to calculate two indicators, one is the distance of the object's movement in the image, and the other is the similarity of the object's background in the image. Finally, these two indicators are combined to determine whether the object is moving relative to the ground.}
		\label{motion}
	\end{figure*}
	
	The fusion of tracking methods involves combining multiple individual tracking methods or models to improve tracking performance or address various challenges. This approach can leverage the strengths of different methods, mitigate the limitations of individual methods, and provide more robust and accurate tracking results.
	
	One common method of tracking method fusion is decision-level fusion \cite{prabhakar2002decision,leangOnlineFusionTrackers2018,tang2023exploring}. This method determines the final tracking result by weighting or voting the outputs of multiple trackers. The weights of each tracker can be dynamically adjusted based on its tracking performance in previous frames, giving greater weight to trackers with better performance.
	
	Another common fusion method is feature-level fusion \cite{lan2014multi,pang2020real, xu2021cross,nie2020mmfn,zhou2023robust}. In this approach, multiple trackers extract feature representations of the target and fuse these features. Fusion can be as simple as averaging or weighted averaging, or it can involve more complex model fusion methods such as convolutional neural networks or support vector machines.
	
	Additionally, there are information fusion-based methods \cite{ding2021progressive,yang2023siammmf} that involve fusing motion information, contextual information, or external prior knowledge. These methods improve tracking performance by combining information from multiple sources. For example, the results of multiple trackers can be fused by incorporating the target's motion model, scene context, or appearance model.
	
	Overall, the fusion of tracking methods offers a way to enhance tracking performance by integrating multiple methods or models, leveraging their strengths, and taking advantage of different sources of information.

	\section{Motion State Judgment}
	\label{Motion Judgment}

	\subsection{Fusion Method}

	In this section, we present a straightforward yet highly effective approach for ascertaining the motion status of tracked objects. Our method is designed to precisely determine whether an object is in motion or stationary throughout the tracking procedure.
	\subsection{Quantifying Object Motion}
	MOT methods are employed to discern the motion trajectories of objects within video sequences. In scenarios involving camera motion, the application of the enhanced correlation coefficient maximization(ECC) \cite{evangelidis2008parametric} model enables robust camera motion correction, allowing us to reliably ascertain the relative inter-frame displacement of objects. This computed displacement serves as a foundational metric for characterizing object motion.
	
	We employ the Mahalanobis distance to calculate the motion distance \(D\), as shown in (1):
	\begin{equation}
		D = \sqrt{(\mathbf{c}_1 - \mathbf{c}_2)^T \mathbf{C}^{-1} (\mathbf{c}_1 - \mathbf{c}_2)}
	\end{equation}
	where
	\begin{align}
		\mathbf{c}_1 &= \left[ \frac{x_1 + \frac{w_1}{2}}{2}, \frac{y_1 + \frac{h_1}{2}}{2} \right] \\
		\mathbf{c}_2 &= \left[ \frac{x_2 + \frac{w_2}{2}}{2}, \frac{y_2 + \frac{h_2}{2}}{2} \right]
	\end{align}
	\begin{equation}
		\mathbf{C} = \begin{bmatrix} \frac{w_{1}^{2}}{4} & 0 \\ 0 & \frac{h_{1}^{2}}{4} \end{bmatrix}
	\end{equation}
	
	Here, \(\mathbf{c}_1\) and \(\mathbf{c}_2\) represent the center coordinates of the two bounding boxes, and \((x_1, y_1, w_1, h_1)\) and \((x_2, y_2, w_2, h_2)\) denote the coordinates and dimensions of the two bounding boxes.
	\begin{figure*}[htbp]
		\centering
		\includegraphics[width=5in]{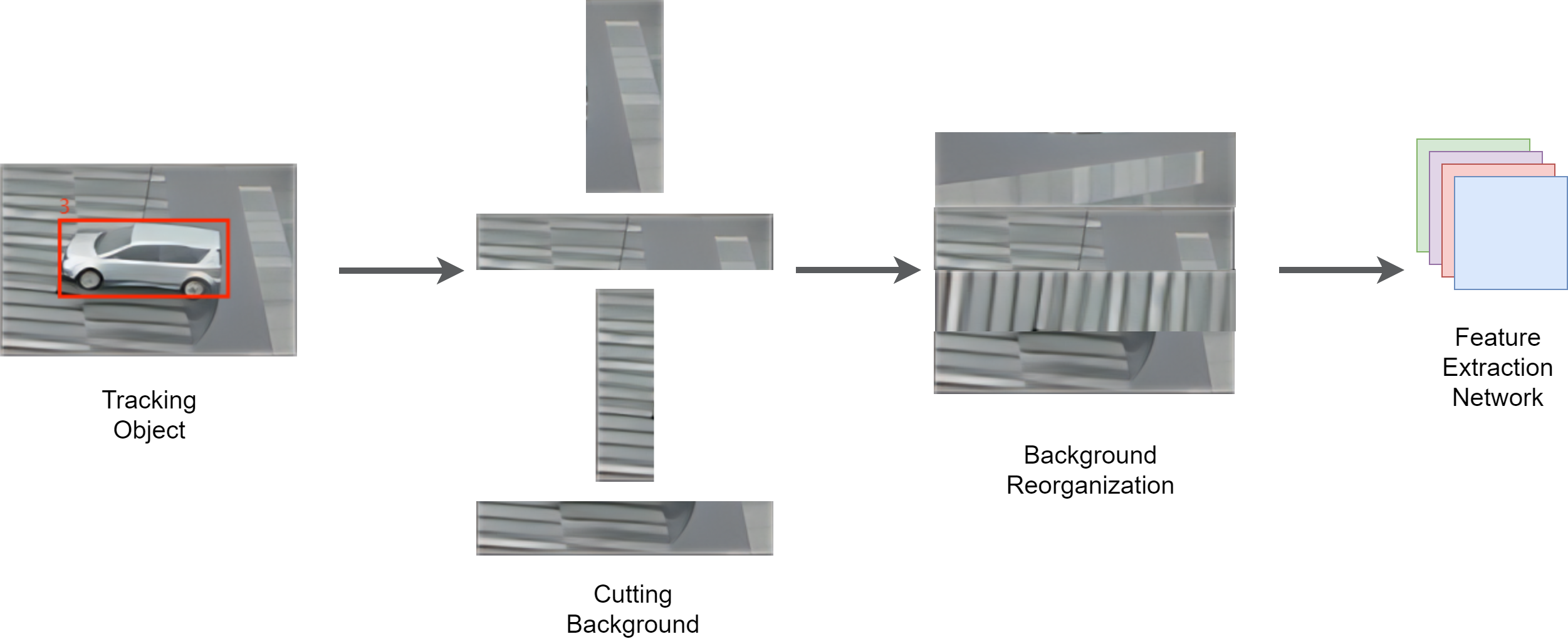}
		\caption{Background feature extraction process. In the context of the feature extraction network, the background images are typically resized to dimensions of 256x192 pixels. Backgrounds from all four orientations are resized to the same dimensions. However, if any side of the background image is within 15 pixels of the image boundary, the portion of the background image close to that boundary is discarded. }
		\label{feature}
	\end{figure*}
	The rationale behind selecting Mahalanobis distance lies in its ability to account for the covariance structure within the data. This becomes particularly crucial when dealing with multidimensional data, as it considers both the variances and covariances among the different dimensions. Notably, we conducted experiments with alternative distance metrics, and Mahalanobis distance demonstrated superior performance in capturing the underlying data structure, especially in handling the intricacies of motion patterns represented by the bounding boxes. This choice significantly contributes to the accuracy and reliability of our motion distance calculations, ensuring robustness in the face of varying data distributions and dimensions.
	\begin{figure*}[htbp]
		\centering
		\includegraphics[width=1\textwidth]{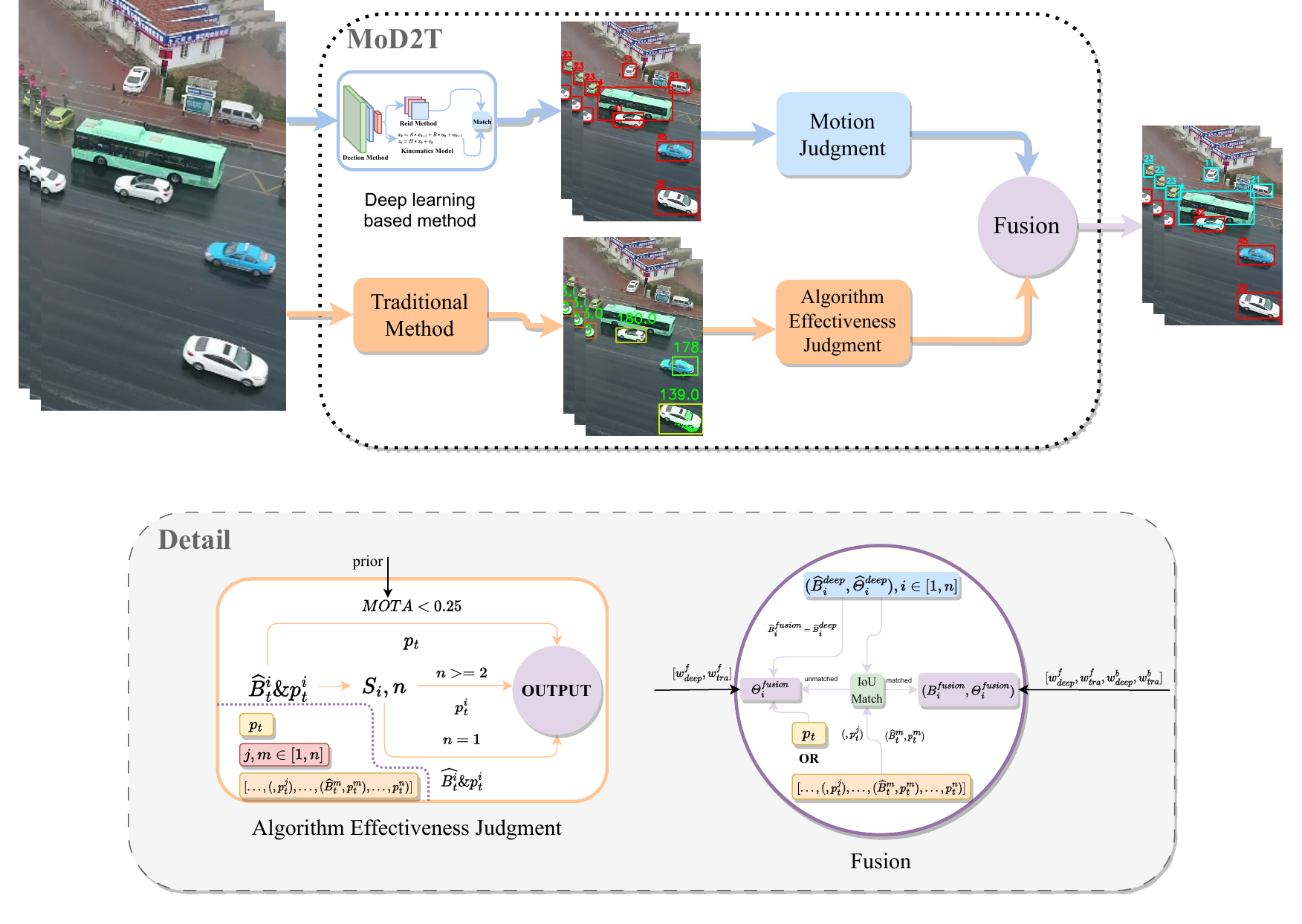}
		\caption{The comprehensive architecture of MoD2T is presented, allowing for flexibility in the integration of various deep learning-based methods. Specifically, this study employs YOLOv8+StrongSORT and YOLOv8+DeepOCSORT as instances of such methods. Figure.\ref{tradition} illustrates the traditional method, while Figure.\ref{motion} focuses on the motion state judgment aspect. Within the algorithmic effectiveness assessment module depicted in the figure, the region enclosed by purple dashed lines signifies multiple potential outputs of this module. The notation $p_t$ within the diagram represents the pixels that traditional algorithms may identify as moving objects.}
		\label{flow}
	\end{figure*}
	\subsection{Background Feature Extraction}
	
	We performed feature extraction on the background of the tracked object. The specific approach involved segmenting the surrounding background and then combining it. We utilized a pre-trained OSNet \cite{zhouOmniScaleFeatureLearning2019} to extract appearance features from this background. The procedure is illustrated in the Figure.\ref{feature}. 
	
	Empirical findings have substantiated the enhancement in accuracy resulting from the incorporation of this procedural step. The discernible rise in MVF1 scores, as illustrated in Tables \ref{KITTI}, \ref{MOT17}, and \ref{UAVDT}, under the condition $\lambda=best$, underscores its efficacy in contrast to cases where $\lambda$ is set to 0. The subsequent section will provide a comprehensive explication of the variable $\lambda$, elucidating its impact on MVF1 scores. The optimal value for $\lambda$, denoted as $best$, is determined through systematic parameter tuning, with specific values tailored for each dataset. For brevity, explicit enumeration of these dataset-specific $best$ values is omitted in this context.
	
	\subsection{Motion State Judgment}
	\label{Motion_Judgment_sub}
	After obtaining the background features and motion distance, we utilize the value $\varTheta$, as shown in (7), to determine whether the object is in motion, when $\varTheta$ is bigger than a threshold, the object is considered stationary.
	\begin{equation}
		\varTheta =\lambda A_a+(1-\lambda )A_m
	\end{equation}
	Here, $A_a$ represents the similarity of the background appearance, which is calculated using the structural similarity (SSIM) index. The specific calculation formula is given by (8). $A_m$ represents the normalized value of the motion distance traveled by the tracked object. $\lambda$ is the proportional coefficient.
	\begin{equation}
		A_a(I_1,I_2)=\frac{(2\mu_{I_1}\mu_{I_2}+r_1)(2\sigma_{I_1I_2}+r_2)}{(\mu _{I_1}^{2}+\mu_{I_2}^{2}+r_1)(\sigma_{I_1}^{2}+\sigma_{I_2}^{2}+r_2)}
	\end{equation}
	\begin{equation}
		A_m=\left. \left\{ \begin{array}{c}
			1-\frac{D}{\beta_d},D\leq \beta_d\\
			0,\mathrm{otherwise}\\
		\end{array} \right. \right. 
	\end{equation}
	Among them,$I_1$ and $I_2$ denote two background feature tensors, with $\mu_{I_1}$ and $\mu_{I_2}$ representing their respective means, $\sigma_{I_1}^{2}$ and $\sigma_{I_2}^{2}$ denoting their variances, and $\sigma_{I_1I_2}$ indicating the covariance between them. Constants $r_1$ and $r_2$ are introduced to prevent a zero denominator, and $\beta_d$ represents the maximum distance a preset stationary object may move within the image.
	
	In our approach, the value of $\beta_d$ is typically empirically set to 0.01 times the smaller dimension (length or width) of the image. This choice ensures a proportional scaling of $\beta_d$ and serves as a small adjustment factor based on the image's dimensions. The rationale behind this selection is to strike a balance by considering image size in computations while maintaining a relatively small value to prevent the introduction of overly significant weights. The choice of using 0.01 as a multiplier is likely an empirical decision aimed at achieving a practical balance, ensuring algorithm effectiveness while minimizing computational demands.

	\section{MoD2T}
	In the previous sections, we discussed the advantages and limitations of both traditional methods and deep learning methods. The objective of this paper is to parallelize and integrate these two approaches to achieve improved results. We propose a model-data-driven motion-static object tracking method (MoD2T) that combines the outputs of traditional method and deep learning method to enhance tracking accuracy and robustness.
	
	The traditional method used in this paper proposes an improvement to the Gaussian model proposed in \cite{yiDetectionMovingObjects2013}, and through experiments, it demonstrates that this method, when combined with the OC-SORT\cite{cao2023observation}, achieves satisfactory results in many environments. For the specific implementation details, please refer to the appendix.	Inspired by the work in \cite{leangOnlineFusionTrackers2018}, which demonstrated the benefits of weighted fusion for multiple tracking methods, we design a fusion scheme for the methods employed in this paper.
	
	The flowchart depicted in Figure \ref{flow} outlines the MoD2T process, where two tracking methods, the traditional method represented as $M^{tra}$ and the deep learning-based method denoted as $M^{deep}$, run simultaneously. For each frame $I_t$, where $t=1,2,...,T$ and $T$ is the total number of frames in the video, both methods generate a series of bounding boxes ($\widehat{B}_{t}^{i}$) and corresponding motion assessment values ($\widehat{\varTheta}_{t}^{i}$), where $i=1,...,n$ and $n$ represents the number of detections in frame $t$. This paper's fusion approach combines their individual outputs, resulting in $\widehat{B}_{t} = (\widehat{B}_{t}^{tra}, \widehat{B}_{t}^{deep})$ and $\widehat{\varTheta}{t} = (\widehat{\varTheta}_{t}^{tra}, \widehat{\varTheta}_{t}^{deep})$. The fused tracking outputs, denoted as $\widehat{B}_{t}^{fusion}$ and $\widehat{\varTheta}_{t}^{fusion}$, are then obtained through this fusion process.
	
	Considering the performance issues associated with traditional methods, our fusion strategy may undergo certain adjustments based on the varying performance of the traditional methods.
	
	Subsequent to the independent processing of image data using two distinct methodologies, we apply the motion classification technique outlined in section \ref{Motion Judgment} to the outcomes derived from the deep learning method. The precise fusion strategy will be contingent upon the efficacy of the traditional approach.
	
	\subsection{Traditional Methods Evaluation}
	\label{Traditional_Evaluation}
	Before proceeding with the evaluation, we establish two key premises. Firstly, if it is known in advance that the traditional method yields a result with an Intersection over MOTA\cite{bernardin2008evaluating} score greater than 0.25 on a similar dataset, then the following evaluation process will be used. Secondly, in the absence of the aforementioned MOTA condition, we check whether the proportion of foreground pixels detected by the traditional method exceeds 0.5 of the entire image size. Only if this condition is met and the MOTA criterion is not satisfied, will we proceed with the evaluation using the following approach. It is imperative to note that if neither of these conditions is met, both criteria imply the inadequacy of traditional methods. The threshold of 0.5 for the proportion of foreground pixels is chosen based on the understanding that values exceeding 0.5 indicate rapid camera motion, rendering traditional methods ineffective. The selection of MOTA as a primary judgment criterion is justified by the potential limitations of traditional methods, which may lead to misunderstandings or overlook the detection of objects in scenarios where motion is not accurately captured.

	Our evaluation is not based on the overall results of the entire image, but rather on evaluating individual tracking results. This is because traditional methods can produce erroneous results when performing morphological operations such as erosion and dilation, leading to two objects being mistakenly identified as one. Therefore, we focus on the connected components within the bounding boxes of each moving object obtained by the traditional method for evaluation.
	
	For each motion pixel $p$, we examine its surrounding neighboring pixels within a small range, typically set to three pixels, inspired by the concept of a 3x3 convolution kernel commonly used in Convolutional Neural Networks (CNN). This approach is akin to the local receptive field employed by CNN, designed to capture local information in images. If additional motion pixels are detected within this specified range, the current pixel is designated as a new connected component and is assigned a unique label. Subsequently, we analyze the resulting labeled image and quantify the number of connected components. This choice of a small range, specifically three pixels, draws inspiration from the effectiveness of 3x3 convolutional kernels in CNN, ensuring a balance between computational efficiency and the model's ability to identify and label connected components of motion pixels.
	
\begin{figure}[htbp]
		\centering
		\includegraphics[width=2in]{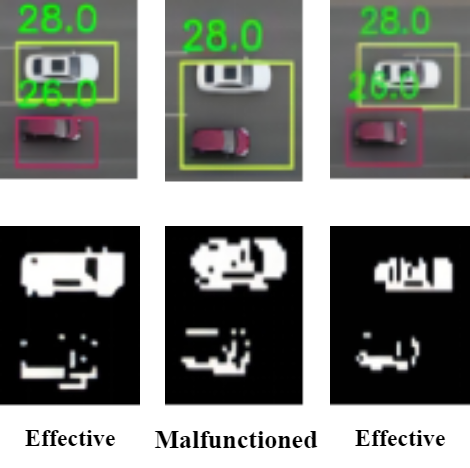}
		\caption{Schematic diagram of whether traditional methods are effective. From the figure, it can be observed that when two objects are close in distance, traditional methods become ineffective. At this point, we should change our approach to fusion.}
		\label{is_eff}
	\end{figure}

	We sum the pixels in these connected components, and if the pixel value exceeds $S_{min}$, we consider it as a valid connected component. The specific expression of this calculation can be represented by the (10): 
	\begin{equation}
		\begin{aligned}
			S&=\sum_{j=1}^n{p^i_j}\\
			n&=\sum_{i=1}^N{[S_i>S_{min}]}\\
		\end{aligned}
	\end{equation}
	When $n\geq 2,$ we consider that the traditional method has not worked properly.
	By calculating and observing the number of connected components obtained by the traditional method, we can assess its performance in handling situations involving close or overlapping objects. This evaluation method allows for a more detailed understanding of the method's performance on individual tracking results and provides an assessment of its effectiveness. One more thing, in this paper, $S_{min}$ generally set to one-fifth of the bbox area. A typical example of the effectiveness of a traditional tracking method is illustrated in Figure. \ref{is_eff}.
	
	\subsection{Fusion Strategies in the Presence of Effective Traditional Methods}
	First, we use Intersection of Union(IoU) to associate and match the output results of the two methods. Then we judge whether the traditional method is effective.
	When the traditional methods work well, we can attempt to fuse both bbox and motion assessment values $\varTheta$. 
	For the tracking results $B^{fusion}_{i},i\,\,\in \,\,\left[ 1,n \right]$ of a single frame, which includes $n$ tracked objects, the fusion is calculated by merging the outputs of both trackers. In this paper, we compute the bbox for each individual object using a weighted average, as shown in (11):
	\begin{equation}
		B_{i}^{fusion}=w_{deep}^{b}*B_{i}^{deep}\,\,+\,\,w_{tra}^{b}*B_{i}^{tra}
	\end{equation}
	Here, the weights denoted as $w_k^b$, where the subscript $k$ signifies distinct methods, are considered as tunable parameters. It is important to emphasize that the weights presented in this study are computed with respect to the performance exhibited by each method on a common dataset. The precise computation of these weight coefficients is delineated in (12):
	\begin{equation}
		\begin{gathered}
			w_{deep}^{b}=\alpha _{k}^{b}\frac{MOTA_{deep}}{MOTA_{tra}+MOTA_{deep}}
			\\
			w_{tra}^{b}=\left( 1-\alpha _{k}^{b} \right) \frac{MOTA_{tra}}{MOTA_{tra}+MOTA_{deep}}
		\end{gathered}
	\end{equation}
	where, $\alpha _{k}^{b}$ is the confidence level, given by humans, and MOTA is a common evaluation indicator for multi-objective tracking methods\cite{bernardin2008evaluating}. The reason for choosing MOTA is that it mainly reflects whether the object has been tracked, and traditional methods often miss detection of the object. Therefore, using this indicator as the standard for calculating fusion parameters is reasonable.
	
	Since deep learning methods track both moving and stationary objects, the tracking results from deep learning often contain more detections compared to traditional methods. In cases where there are no corresponding detections from the traditional method for the deep learning results, we directly adopt the results from the deep learning method. In other words, we set $w_{deep}^{b}=1,w_{tra}^{b}=0$ in this cases. This ensures that the deep learning results are considered as the primary source of tracking information when there is no corresponding result from the traditional methods.
	
	The value $\varTheta$, which indicates whether an object is in motion or not, is also calculated using a weighted average in this paper, as shown in (13). The calculation of $\varTheta^{deep}$ is defined in (7). For $\varTheta_{i}^{tra}$, when there is no corresponding detection from the traditional method for the deep learning results, it is set to 1; otherwise, it is set to 0. Meanwhile, if the traditional method does not work properly, its value will be shown in equation (16)
	
	\begin{equation}
		\varTheta_{i}^{fusion}=\left. w_{deep}^{f}*\varTheta_{i}^{deep}\,\,+\,\,w_{tra}^{f}*\varTheta_{i}^{tra}\,\, \right. 
	\end{equation}
	Subscript $i$ means the ID of different tracking objects.
	
	Similar to the weights for bbox, the weights for $\varTheta$ are also determined based on the performance of each method on the same dataset. The weight coefficients can be calculated as shown in (14), 
	\begin{equation}
		\begin{gathered}
			w_{deep}^{f}=\frac{MVF1_{deep}}{MVF1_{tra}+MVF1_{deep}}
			\\
			w_{tra}^{f}= \frac{MVF1_{tra}}{MVF1_{tra}+MVF1_{deep}}
		\end{gathered}
	\end{equation}
	where the $\text{MVF1}$ is an indicator used to measure whether the judgment of a moving object is accurate, and the specific calculation method is shown in (15). The $\text{MVF1}_{tra}$ to be 1.If traditional methods can still track performance ($\text{MOTA}>=0.25 $), then $\text{MVF1}_{tra}$ is 1, otherwise it is 0.2. The reason for setting these two values is based on experimental results.
	
	\subsection{Fusion Strategy in the Presence of Malfunctioning Traditional Methods}
	\label{Fusion_strategy}
	When the performance of traditional method is inadequate, meaning it performs poorly on pre-prepared test sets or when the judgment criteria proposed in this paper deem their performance to be subpar, only the value $\varTheta$ is fused as (13). The calculation for $\varTheta_{i}^{fusion}$ and $C_{i}^{deep}$ remains unchanged. However, the calculation for $\varTheta_{i}^{tra}$ will be calculated as shown in (16). The bbox will directly adopt the results from the deep learning-based MOT method.
	
	\begin{equation}
		\varTheta_{i}^{tra}=\begin{cases}
			\frac{p_i}{w*h*R},&		\frac{p_i}{w*h}<R\\
			1,&		\mathrm{otherwise}\\
		\end{cases}
	\end{equation}
	Where $w$ and $h$ represent the width and height of the bounding box, respectively, and $p_i$ denotes the number of pixels in the bounding box that correspond to the motion detection results from the traditional method. $R$ is a correction value, which is experimentally determined to be 0.2. This value is consistent with the determination of the S value in section \ref{Traditional_Evaluation}.

	\subsection{Object Motion Metrics: Motion State Validation F1 Score}
	
	Recognizing the absence of a specialized metric for evaluating the accuracy of models in predicting object motion states, we drew inspiration from the F1 score to introduce the $MVF1$ metric as an innovative evaluation criterion.
	
	\begin{equation}
		\begin{gathered}
			MVF1=2\cdot \frac{MVP\cdot MVR}{MVP+MVR}
			\\
			MVP=\frac{MVTP}{MVTP+MVFP}
			\\
			MVR=\frac{MVTP}{MVTP+BF \cdot MVFN}
		\end{gathered}
	\end{equation}
	\noindent Here, $MVTP$ signifies the count of samples accurately classified as either moving or stationary objects, $MVFP$ denotes the count of samples inaccurately labeled as moving objects when they are stationary or as stationary objects when they are in motion, and $MVFN$ represents the count of samples absent from the ground truth. $BF$ stands for the balance factor, a parameter adjustable to enhance adaptability.
	
	The introduction of $BF$ is motivated by the metric's primary purpose of measuring the accuracy in determining the motion state of objects. Ideally, the impact of tracking or detection performance should be disregarded. Nevertheless, given that motion-state determination relies on tracking, this metric retains a certain influence from the detection process.
	
	\begin{equation}
		BF = \frac{1}{2} \cdot (\text{precision} + \text{recall})
	\end{equation}
	where \text{precision} and \text{recall} represent the precision and recall of the object detection. Consequently, when precision and recall are balanced, $BF$ converges to 0.5, mitigating the impact of MVFN. As precision or recall deviates from equilibrium, $BF$ adjusts dynamically, taking into account undetected objects based on the overall performance of object detection. As for why detection metrics are used as balance parameters, it is mainly because the tracking performance and detection performance of almost all current tracking algorithms are closely related.
	
	%	Certainly, to elucidate the outcomes of the Bayes factor (BF) parameter, this study also establishes a counterpart without the inclusion of the BF parameter, denoted as $MVF1^w$.
	%
	%	\begin{equation}
		%	\begin{gathered}
			%		MVF1^w=2\cdot \frac{MVP\cdot MVR}{MVP+MVR^w}
			%		\\
			%		MVP=\frac{MVTP}{MVTP+MVFP}
			%		\\
			%		MVR^w=\frac{MVTP}{MVTP+MVFN}
			%	\end{gathered}
		%	\end{equation}

	\section{Experiment}
	In this section, we conduct extensive experiments to evaluate the performance of MoD2T on several widely used datasets and compare it with several state-of-the-art MOT methods. We also analyze the impact of various components in MoD2T and provide insights into its effectiveness and limitations.
	Since we have not found any work that can fully compare with the work in this paper, this article did not set a baseline for comparison. We found that the work closest to the core work of this paper is various tasks for image change detection such as CDnet-2014\cite{wangCDnet2014Expanded2014}, which include\cite{rahmonMotionUNetMulticue2021,limLearningMultiscaleFeatures2020} , but none of these works are suitable for comparison with this paper.
	
	\subsection{Setting}
	\textbf{Datasets.} Currently, a void exists in the availability of multi-object tracking datasets that incorporate annotations about the dynamic state of objects. Nevertheless, a recourse is available through repurposing datasets originally crafted for autonomous driving, such as the KITTI \cite{Geiger2012CVPR}. This dataset includes precise 3D coordinates of objects, thereby facilitating the determination of object motion status within the tracking context.
	
	The KITTI dataset \cite{Geiger2012CVPR} has achieved wide adoption within the autonomous driving and computer vision research communities. It encompasses data collected from diverse sensors like cameras and LiDAR mounted on vehicles. The dataset presents an array of scenes annotated with object bounding boxes and motion trajectories for various entities, including vehicles and pedestrians, thereby rendering it a fitting choice for the evaluation of multi-object tracking methodologies.
	
	Moreover, our investigation extends to selected scenarios from the UAVDT \cite{du2018unmanned} and MOT17 \cite{milanMOT16BenchmarkMultiObject2016} datasets. The UAVDT dataset is meticulously tailored for video analysis of unmanned aerial vehicles, featuring captured video sequences across a spectrum of scenarios. On the other hand, the MOT17 dataset emphasizes multi-object tracking pertaining to pedestrians, offering video footage captured from different vantage points along with annotations detailing the motion trajectories of multiple objects.
	
	To underscore the versatility and efficacy of our method, we subject these UAVDT and MOT17 scenes to supplementary evaluation, bolstered by augmented annotations. These curated scenes introduce heightened complexity and intricate tracking scenarios, thereby affording a comprehensive assessment of the method's resilience and adaptability. By subjecting our method to scrutiny across diverse datasets, we attain a holistic comprehension of its performance in multi-object tracking tasks and corroborate its efficacy across a plethora of scenarios.
	
	Our assessment strategy was systematic, entailing an exhaustive performance analysis conducted half of KITTI's training set for training and estimating some parameters, and use the other half for testing. Given the need for supplemental annotations concerning object motion attributes within the MOT17 and UAVDT datasets, the entirety of these datasets was not embraced in this inquiry. Instead, our evaluation was concentrated on representative scenes, specifically MOT17 sequences (03, 07) and UAVDT sequences (M101, M202, M204, M402, M601, M703, M802).
	
	The outcomes of these exacting evaluations have been scrupulously collated in the forthcoming table. In a bid to furnish a comprehensive insight into the effectiveness and prowess of our proposed approach, we partitioned the KITTI dataset into two discrete segments for the purpose of testing. One segment encompasses sequences (0001,0017-0024), characterized by static or slowly moving cameras, while the other segment encapsulates scenarios involving swifter camera motions and accelerated object movements.
	
	One more thing, although using the test datasets for calculating the fusion weights, it is anticipated that this has minimal impact on the results. Nevertheless, in order to ensure the fairness of our experiments, this paper also employed additional datasets for a rough estimation of these weights. Specifically, for KITTI, the additional data encompassed the entire training set, which was further categorized based on the speed of motion. For MOT17, the selected datasets were 04 and 10, and for UAVDT, it included M203, M403, and M701. It is important to emphasize that although these datasets underwent additional annotation and can be utilized for computing tracking performance metrics and MVF1 scores, they were not used in the method's performance testing in this study.

	\textbf{Metrics.} We use the metrics MOTA, IDF1,AssA and HOTA to evaluate tracking performance\cite{bernardin2008evaluating,ristaniPerformanceMeasuresData2016,Luiten2020IJCV}. MOTA primarily measures the completeness and robustness of tracking, IDF1 focuses on the accuracy and precision of targets, while HOTA comprehensively considers multiple aspects of performance, including both target and trajectory-level accuracy. When comparing methods and evaluating performance, these metrics can assist researchers in gaining a comprehensive understanding of the strengths, weaknesses, and applicability of the methods.
	
	Additionally, we use the MVF1 to measure the accuracy of object motion state predictions. The calculation method for the MVF1 is referenced in (15).

	\textbf{Implementation Details.}  We utilize OSNet\cite{zhouOmniScaleFeatureLearning2019} as the network for extracting background appearance features. For testing on the UAVDT and KITTI , we employ osnet pre-trained on VeRi\cite{liu2016deep}, while for evaluating the MOT17 , we use weights pre-trained on Market1501\cite{zheng2015person}. Additionally, to conserve computational resources, the same osnet network is used in the deep learning-based MOT method adopted in this study. The criterion used to determine correctness is the Intersection over Union (IoU). A prediction is considered correct when the IoU between the method's prediction and the ground truth is greater than 0.5. The training method of YOLOv8 in MOT17 is similar to \cite{zhangByteTrackMultiobjectTracking2022}. The tracker used in traditional methods is OC-SORT\cite{cao2023observation}. Specifically, for the KITTI dataset, this paper divides each sequence in its training set into two equally large parts, with the first part used for training and the second part used for testing.
	\subsection{Reasonability analysis of MVF1}
	
	As mentioned earlier, we propose an indicator for evaluating the accuracy of the model's prediction of object motion states. The definition of FN determines that this indicator is actually not very related to the main content we want to evaluate. Here, to demonstrate the rationality of the FN definition and elucidate the utility of BF parameter, we selected two sequences from the UAVDT dataset, M203 and M403, for testing, totaling approximately 2000 frames.. We examined the impact of varying detection performance on the MVF1 metric. Different detection accuracies were approximated by manually removing results of the same identity throughout the entire sequence.
	\begin{figure}[htbp]
		\centering
		\includegraphics[width=3.5in]{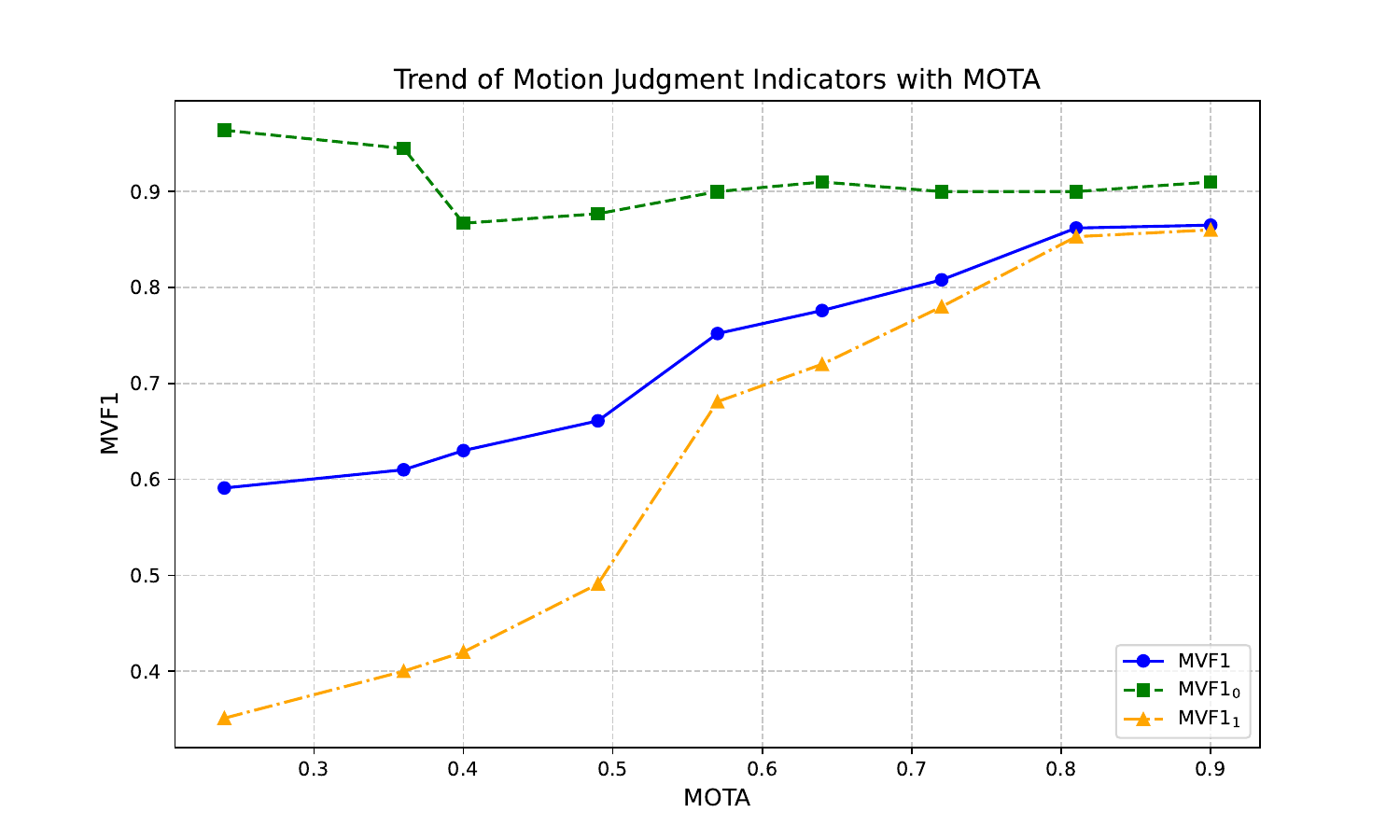}
		\caption{
			The fluctuations observed in MVF1 are attributed to variations in tracking performance trends. The blue curve signifies alterations introduced by adjusting the parameters related to the BF, while the green curve corresponds to the scenario where the BF parameter is set to 1, and the red curve signifies its configuration with a BF parameter value of 0.
		}
		\label{MVF1_trend}
	\end{figure}
	
	As depicted in the Figure.\ref{MVF1_trend}, the dynamic changes observed in MVF1, driven by varying levels of MOTA, offer valuable insights into the nuanced impact of BF parameter configurations on the model's ability to assess the relative motion states of tracked objects with respect to the ground. Furthermore, it is noteworthy that distinct MOTA values were generated by randomly removing entire sequences with identical ID numbers.
	
	Examining the blue curve, representing MVF1 values under varying BF parameters, a clear positive correlation emerges with increasing MOTA levels. This trend emphasizes the utility of strategic BF parameter adjustments in enhancing the model's capacity to accurately discern the motion states of tracked objects relative to the ground. Notably, the performance improvement is most pronounced at higher MOTA values, underscoring the synergistic relationship between MOTA and BF parameter optimization.
	
	Conversely, the green curve, denoting the scenario where the BF parameter is set to 1, suggests a nuanced trade-off. While initial MVF1 values exhibit promise, a subsequent decline in performance is observed as MOTA increases. This nuanced relationship underscores the importance of balancing BF parameter settings, as extreme values may compromise tracking accuracy.
	
	The orange curve, corresponding to a BF parameter value of 0, reveals intriguing stability in MVF1 values across varying MOTA levels. This finding indicates that, in specific contexts, the deliberate exclusion of background filtering may be a viable choice, particularly when minimizing false negatives is a critical consideration.
	
	In essence, the result demonstrates that judicious BF parameter settings significantly contribute to the enhanced evaluative capacity of MVF1, offering a more nuanced assessment of the model's proficiency in discerning the motion states of tracked objects in relation to the ground. The observed trends highlight the need for a context-aware approach to BF parameter tuning, emphasizing the delicate balance required for optimal tracking performance.

	\subsection{Motion State Judgment Results}
	
	\begin{table*}[htbp]
		\centering
		\caption{MoD2T Motion State Judgment Performance on KITTI\label{KITTI_state}}
		\begin{tabular}{l|c|c|c|c|c|c}
			\toprule[1pt]
			~ & \multicolumn{3}{c|}{Camera stationary or moving slowly} & \multicolumn{3}{c}{Camera moving fast} \\ \hline
			Method & $\text{MVF1}$ & $\text{MVF1}_0$ & $\text{MVF1}_1$ & $\text{MVF1}$& $\text{MVF1}_0$& $\text{MVF1}_1$ \\ \hline
			StrongSORT\cite{duStrongSORTMakeDeepSORT2023}+MJ($\lambda=0$) & 0.392 & 0.427 & 0.385 & 0.221& 0.258& 0.209 \\
			StrongSORT+MJ($\lambda=best$) & 0.474 & 0.503 & 0.422 & 0.219& 0.261& 0.210 \\
			StrongSORT+MJ($\lambda=1$) & 0.426 & 0.468 & 0.420 &0.111& 0.158& 0.097 \\
			DeepOCSORT\cite{maggiolinoDeepOCSORTMultiPedestrian2023}+MJ($\lambda=0$) & 0.394 & 0.425 & 0.388 & 0.213& 0.252& 0.206 \\
			DeepOCSORT+MJ($\lambda=best$) &  0.472 & 0.501 & 0.425 & 0.220& 0.262& 0.213\\
			DeepOCSORT+MJ($\lambda=1$) & 0.425 & 0.469 & 0.410 & 0.113& 0.157& 0.099\\
			MoD2T(StrongSORT) & \textbf{0.775} & 0.821 & 0.763 & 0.230& 0.239& 0.228 \\
			MoD2T(DeepOCSORT) & 0.774 & \textbf{0.822} & \textbf{0.764} & 0.231& 0.237& 0.227 \\
			\bottomrule[1pt]
		\end{tabular}
	\end{table*}
	
	\begin{table}[htbp]
		\centering
		\caption{MoD2T Motion State Judgment Performance on MOT17\label{MOT17_state}}
		\begin{tabular}{l|c|c|c}
			\toprule[1pt]
			Method & $\text{MVF1}$ & $\text{MVF1}_0$ & $\text{MVF1}_1$\\ \hline 
			StrongSORT+MJ($\lambda=0$) & 0.397 & 0.436 & 0.392  \\
			StrongSORT+MJ($\lambda=best$) & 0.418 & 0.445 & 0.387 \\
			StrongSORT+MJ($\lambda=1$) & 0.352 & 0.382 & 0.331 \\
			DeepOCSORT+MJ($\lambda=0$) & 0.396 & 0.431 & 0.389 \\
			DeepOCSORT+MJ($\lambda=best$) & 0.428 & 0.457 & 0.398 \\
			DeepOCSORT+MJ($\lambda=1$) & 0.354 & 0.386 & 0.337  \\
			MoD2T(StrongSORT) & 0.518 & \textbf{0.583} & \textbf{0.483}  \\
			MoD2T(DeepOCSORT) & \textbf{0.521} & \textbf{0.583} & 0.481 \\
			\bottomrule[1pt]
		\end{tabular}
	\end{table}
	
	\begin{table}[htbp]
		\centering
		\caption{MoD2T Motion State Judgment Performance on UAVDT\label{UAVDT_state}}
		\begin{tabular}{l|l|c|c}
			\toprule[1pt]
			Method & MVF1 & $\text{MVF1}_0$ & $\text{MVF1}_1$ \\ \hline 
			StrongSORT+MJ($\lambda=0$) & 0.414 & 0.452& 0.408  \\
			StrongSORT+MJ($\lambda=best$) & 0.527 & 0.563& 0.522 \\
			StrongSORT+MJ($\lambda=1$) & 0.385 & 0.401& 0.381  \\
			DeepOCSORT+MJ($\lambda=0$) & 0.438 & 0.461& 0.401 \\
			DeepOCSORT+MJ($\lambda=best$) & 0.525 & 0.561& 0.520 \\
			DeepOCSORT+MJ($\lambda=1$) & 0.390 & 0.413& 0.383  \\
			MoD2T(StrongSORT) & 0.825 & \textbf{0.862}& 0.820  \\
			MoD2T(DeepOCSORT) & \textbf{0.827} & 0.861& \textbf{0.824} \\
			\bottomrule[1pt]
		\end{tabular}
	\end{table}

	The specific outcomes are delineated in Tables \ref{KITTI_state}, \ref{MOT17_state}, and \ref{UAVDT_state}. In these tables, the term "MJ" denotes motion state judgment, $\text{MVF1}_0$ signifies a background filtering (BF) coefficient set to 0, and $\text{MVF1}_1$ denotes a BF coefficient set to 1. $\lambda$ is a proportional coefficient, and its specific meaning can be referred to \ref{Motion_Judgment_sub}. Meanwhile, as this paper should be the first to propose an method for judging the motion state of objects in MOT scenarios, we have not found any other work that can be compared with the algorithm proposed in this paper.
	
	Directing attention to the KITTI dataset, encompassing scenes characterized by diverse camera motions, MoD2T demonstrates noteworthy enhancements in the MVF1 score, particularly in scenarios involving slow camera motion or a stationary camera. The method excels in accurately discerning static from moving objects. However, despite these advancements, the overall tracking performance across all methods remains consistent, indicating that the improvements predominantly contribute to the discrimination of static or slowly moving objects, rather than constituting a substantial enhancement to tracking capabilities. Notably, MoD2T's capacity to discern object motion diminishes in scenarios characterized by rapid camera motion.
	
	In the UAVDT dataset, MoD2T exhibits a significant elevation in the MVF1 score compared to deep learning-based multiple object tracking (MOT) methods, such as StrongSORT and DeepOCSORT. Analogous to the KITTI dataset, the tracking performance remains relatively unaffected, underscoring MoD2T's proficiency in detecting static or slowly moving objects without substantial gains in overall tracking capabilities.
	
	Similarly, in the MOT17 dataset, analogous trends emerge. MoD2T's efficacy in determining object motion is notably superior in scenarios with slow camera motion or a stationary camera (e.g., MOT17-03), in contrast to situations featuring rapid camera motion (e.g., MOT17-07). Moreover, when cross-referencing datasets, such as UAVDT and KITTI, wherein camera motion is minimal, it becomes apparent that discerning human motion is considerably more challenging than detecting vehicle movement.
	
	The similarities in results between DeepOCSORT and StrongSORT highlight their comparable tracking performance in various motion state scenarios, assessed through the MVF1 metric and its derivatives ($\text{MVF1}_0$ and $\text{MVF1}_1$) in dynamic camera environments.
	
	Across different datasets (KITTI, MOT17, and UAVDT), both methods consistently exhibit similar trends in MVF1, $\text{MVF1}_0$ and $\text{MVF1}_1$, indicating a sustained level of agreement in motion state judgment performance.
	
	This alignment in outcomes is likely attributable to shared foundational principles or algorithmic components, resulting in similar responses to diverse motion state scenarios. Both methods may employ comparable strategies to address challenges posed by stationary or slowly moving objects versus swiftly moving entities.
	
	Minor disparities observed in specific instances may stem from dataset-specific attributes or variations in parameters, such as the tuning parameter $\lambda$ in motion state judgment module. Nevertheless, the overall trend of closely aligned performance suggests a robust and consistent behavior exhibited by both tracking approaches across diverse scenarios.
	
	In summary, the consistent alignment of results across various datasets emphasizes the comparable effectiveness of DeepOCSORT and StrongSORT in motion state judgment. Despite potential differences in their underlying architectures, both methods consistently deliver similar tracking outcomes, showcasing their proficiency in addressing diverse motion scenarios encountered in real-world tracking applications.

	\begin{table*}[htbp]
		\centering
		\caption{MoD2T Tracking Performance on KITTI\label{KITTI}}
		\begin{tabular}{l|l|l|l|l|l|l|l|l}
			\toprule[1pt]
			~ & \multicolumn{4}{c|}{Pedestrian} & \multicolumn{4}{c}{ Car} \\ \hline
			Method & HOTA & IDF1 & MOTA & AssA & HOTA & IDF1 & MOTA & AssA \\ \hline
			StrongSORT\cite{duStrongSORTMakeDeepSORT2023} & 0.812 & 0.876 & 0.877 & 0.838 & 0.852 & 0.915 & 0.915 & 0.858 \\
			DeepOCSORT\cite{maggiolinoDeepOCSORTMultiPedestrian2023}& 0.822 & 0.874 & 0.886 & 0.828 & 0.859 & 0.921 & 0.922 & 0.861 \\
			MoD2T(StrongSORT) & 0.811 & 0.876 & 0.877 & 0.837  & 0.852 & 0.915 & 0.915 & 0.858 \\
			MoD2T(DeepOCSORT) & 0.822 & 0.874 & 0.886 & 0.828 & 0.859 & 0.921 & 0.922 & 0.861 \\
			\bottomrule[1pt]
		\end{tabular}
	\end{table*}
	
	\begin{table}[htbp]
		\centering
		\caption{MoD2T Tracking Performance on MOT17\label{MOT17}}
		\begin{tabular}{l|l|l|l|l}
			\toprule[1pt]
			Method & HOTA & IDF1 & MOTA & AssA \\ \hline
			StrongSORT & 0.682 & 0.821 & 0.825 & 0.682 \\
			DeepOCSORT & 0.714 & 0.833 & 0.825 & 0.726 \\
			MoD2T(StrongSORT) & 0.681 & 0.821 & 0.823 & 0.682 \\
			MoD2T(DeepOCSORT) & 0.714 & 0.833 & 0.825 & 0.726 \\
			\bottomrule[1pt]
		\end{tabular}
	\end{table}
	
	\begin{table}[htbp]
		\centering
		\caption{MoD2T Tracking Performance on UAVDT\label{UAVDT}}
		\begin{tabular}{l|l|l|l|l}
			\toprule[1pt]
			Method & HOTA & IDF1 & MOTA & AssA \\ \hline
			StrongSORT & 0.812 & 0.931 & 0.932 & 0.815 \\
			DeepOCSORT & 0.828 & 0.943 & 0.945 & 0.820 \\
			MoD2T(StrongSORT) & 0.812 & 0.931 & 0.932 & 0.896 \\
			MoD2T(DeepOCSORT) & 0.828 & 0.943 & 0.945 & 0.820 \\
			\bottomrule[1pt]
		\end{tabular}
	\end{table}

	\subsection{Tracking Results}
	MoD2T's tracking performance is meticulously evaluated across KITTI, MOT17, and UAVDT datasets, as detailed in Tables \ref{KITTI}, \ref{MOT17}, and \ref{UAVDT}. The assessment includes key metrics such as HOTA, IDF1, MOTA, and AssA.
	
	In the KITTI, MoD2T, synergized with both StrongSORT and DeepOCSORT, exhibits tracking metrics on par with individual methods. Crucially, HOTA, IDF1, MOTA, and AssA scores remain consistent, affirming that MoD2T seamlessly integrates without compromising the intrinsic tracking performance of the base methods. This emphasizes MoD2T's compatibility, preserving the effectiveness of existing models.
	
	The MOT17 reveals a parallel trend, where MoD2T sustains the performance levels of StrongSORT and DeepOCSORT. HOTA, IDF1, MOTA, and AssA scores for MoD2T align closely with standalone methods, corroborating that MoD2T augmentation does not impede the underlying models' tracking proficiency.
	
	UAVDT dataset evaluation further emphasizes MoD2T's consistent trend in preserving and enhancing tracking performance. Particularly, HOTA, IDF1, MOTA, and AssA scores for MoD2T surpass standalone methods in the AssA metric, suggesting MoD2T's potential to improve tracking accuracy in specific scenarios.
	
	In essence, the results highlight that MoD2T, in collaboration with StrongSORT and DeepOCSORT, maintains the robustness of individual tracking models while substantially enhancing tracking performance. The introduced framework ensures that MoD2T incorporation does not degrade the intrinsic tracking capabilities of the base models. This reinforces MoD2T's compatibility and efficacy in real-world tracking applications, positioning it as a valuable enhancement for multi-object tracking tasks.
	
	\begin{figure*}[htbp]
		\centering
		\includegraphics[width=1\textwidth]{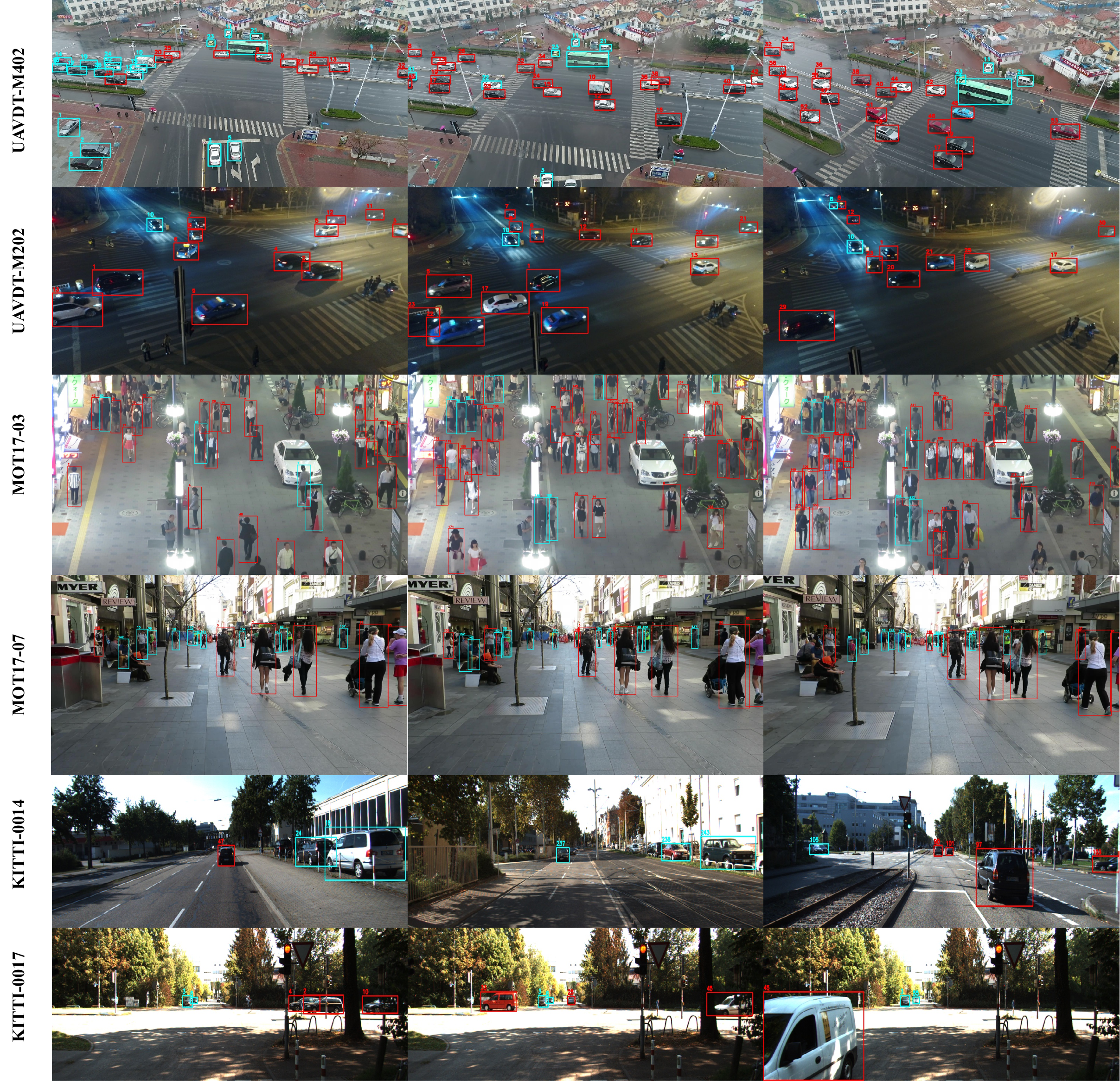}% 笔记本用_1，台式用大的
		\caption{Visualize the sample results of MoD2T on the KITTI and some scenarios of MOT17 and UAVDT. The red box represents moving objects, and teal represents stationary objects. From here, it can be clearly seen that the function of MoD2T to judge the motion status of objects performs very well when the camera is not moving vigorously.}
		\label{result}
	\end{figure*}
	\subsection{Traditional Method Results }
	\begin{table}[htbp]
		\centering
		\caption{Traditional Method Performance on Different Datasets.The correct IoU threshold is 0.5\label{Traditional-performance_0.5}}
		\begin{tabular}{l|c|c|c|c}
			\hline
			Dataset & IDF1 & HOTA & MOTA & AssA \\
			\hline
			KITTI\cite{Geiger2012CVPR}(Moving Fast)  & 0.013 & 0.010 & 0.012 & 0.009 \\
			KITTI(Other)  & 0.063 & 0.019 & 0.063 & 0.012 \\
			MOT17\cite{milanMOT16BenchmarkMultiObject2016} & 0.043 & 0.024 & 0.042 & 0.025 \\
			UAVDT\cite{du2018unmanned} & 0.068 & 0.043 & 0.069 & 0.044 \\
			\hline
		\end{tabular}
	\end{table}
	
	\begin{table}[htbp]
		\centering
		\caption{Traditional Method Performance on Different Datasets.The correct IoU threshold is 0.25\label{Traditional-performance_0.25}}
		\begin{tabular}{l|c|c|c|c}
			\hline
			Dataset & IDF1 & HOTA & MOTA & AssA \\
			\hline
			KITTI\cite{Geiger2012CVPR}(Moving Fast)  & 0.015 & 0.011 & 0.015 & 0.013 \\
			KITTI(Other)  & 0.543 & 0.439 & 0.542 & 0.432 \\
			MOT17\cite{milanMOT16BenchmarkMultiObject2016} & 0.361 & 0.262 & 0.361 & 0.365 \\
			UAVDT\cite{du2018unmanned} & 0.454 & 0.362 & 0.464 & 0.361 \\
			\hline
		\end{tabular}
	\end{table}
	Table \ref{Traditional-performance_0.5} presents the performance of traditional methods on different datasets, with an IoU threshold of 0.5, specifically focusing on correctly tracking moving objects. The results reveal that, at an IoU threshold of 0.5, the performance of traditional methods is generally unsatisfactory across all datasets. Particularly challenging is the KITTI , which involves intense camera motion.
	
	When the IoU threshold is relaxed to 0.25, as shown in Table \ref{Traditional-performance_0.25}, the proposed method achieves marginally better results in most scenarios. This improvement suggests that, for traditional methods, they can often confirm the presence of a moving object in many cases, but their accuracy in precisely determining the object's boundaries is limited.
	
	However, it becomes evident that when dealing with fast camera motion in the KITTI , the performance of traditional algorithms still falls short. This is also one of the reasons why MoD2T underperforms in such scenarios. Therefore, further research and refinement are necessary to enhance the method's robustness in fast-paced environments, or alternatively, consider adopting more robust approaches.
	
	In summary, traditional methods exhibit limited accuracy when evaluating moving objects at an IoU threshold of 0.5. The proposed method shows some improvement when the IoU threshold is relaxed, but it still faces challenges, particularly in scenarios with fast camera motion, where the performance of traditional algorithms also struggles. To address this issue, continuous efforts should be made to enhance the method's robustness in fast-motion environments, or explore alternative, more resilient methodologies. Meanwhile, In comparison to Tables \ref{KITTI}, \ref{MOT17}, and \ref{UAVDT}, MoD2T demonstrates significant advantages in tracking performance over traditional method. 
	
	\subsection{Analysis of Some Parameters}
	This paper has meticulously expounded on the pivotal fusion parameter configuration, elucidating its intricacies in a preceding section. However, certain parameter settings, such as the parameter denoted as $S_{min}$ in section \ref{Traditional_Evaluation}, are derived from experiential considerations. To substantiate the rationale behind these parameter choices, supplementary experimental analyses have been conducted. The objective is to furnish a rational explanation for the specified parameter values, shedding light on the nuanced decisions made in their selection within the context of this research endeavor.
	
	Regarding $S_{min}$, its configuration in Section \ref{Fusion_strategy} mirrors that of $R$ since both parameters signify the ratio of motion pixels to the entire box pixels. Notably, the ratio $S_{min}/R$ precisely corresponds to the size of bbox. So these two parameters are analyzed together. 
	
	To illustrate the rationality of our values. We tested the effectiveness of different R values on the sequence of analyzing the rationality of MVF1 design, and the results are shown in the table below.
	
	\begin{table}[htbp]
		\centering
		\caption{R sensitivity analysis\label{R_or_S}}
		\begin{tabular}{l|c|c|c}
			\toprule[1pt]
			$R$ & $\text{MVF1}$ & $\text{MVF1}_0$ & $\text{MVF1}_1$\\ \hline 
			0.1 & 0.828 & 0.852 & 0.819   \\
			0.15 & 0.844 & 0.863 & 0.839  \\
			0.2 & 0.865 & 0.901 & 0.860 \\
			0.3 & 0.805 & 0.874 & 0.796  \\
			0.5 & 0.687 & 0.727 & 0.674  \\
			0.7 & 0.565 & 0.601 & 0.560   \\
			\bottomrule[1pt]
		\end{tabular}
	\end{table}
	
	The choice of R = 0.2 is justified by a careful sensitivity analysis, as depicted in Table \ref{R_or_S}. A smaller value of R, such as 0.1 or 0.15, makes the MoD2T overly sensitive, leading to the risk of misclassifying static objects as moving. Conversely, a larger R, for instance, 0.3, 0.5, or 0.7, tends to result in reduced sensitivity, potentially causing dynamic objects to be erroneously identified as stationary. Therefore, the selection of R = 0.2 strikes a balance, ensuring optimal sensitivity for motion detection while minimizing the likelihood of false positives or negatives in distinguishing object motion states.

	In Figure \ref{flow}, a notable branch exhibits a MOTA score below 0.25. This specific threshold is selected based on the empirical observation that when the MOTA of the traditional method falls below 0.25, it strongly indicates suboptimal performance on sequences of this nature. In such instances, the traditional method proves ineffective. It is essential to clarify that the value of 0.25 serves as a pragmatic threshold, lacking inherent significance beyond being a chosen reference point for assessing traditional method performance on specific sequences.

	\subsection{Limitations}
	
	MoD2T primarily excels when dealing with minor camera motion, particularly in scenarios involving rotary-wing UAVs. It effectively distinguishes between stationary and slowly moving objects, resulting in higher MVF1 scores and improved detection capabilities. However, its performance is constrained when the camera experiences significant motion, especially in accurately assessing the motion of mobile objects. This limitation, sensitivity to camera motion, significantly impacts its effectiveness in settings with frequent camera motion, like dynamic outdoor environments or rapid camera movements. Additionally, MoD2T's performance varies across different object categories, particularly in assessing motion for vehicles and pedestrians, showing significant disparities. This suggests potential biases in specific scenarios.
	
	Furthermore, MoD2T's accuracy in motion assessment is contingent on the quality of the tracking methodology employed. Generally, better tracking methods lead to higher MVF1. MoD2T shines in scenarios with a stationary camera or minor camera motion, proficiently distinguishing stationary from slowly moving objects. However, in contexts with more intense camera motion, especially in assessing mobile object motion, its performance is limited. There are also performance differences across object categories, with better motion assessment for vehicles compared to pedestrians. Future research should focus on addressing these limitations by enhancing MoD2T's ability to handle pronounced camera motion and improving motion assessment accuracy across all object categories.
	
	\section{Conclusion}

In this paper, we introduce a novel benchmark in the domain of MOT: motion state. Subsequently, addressing this benchmark, we present MoD2T, a pioneering approach to MOT in video analysis. MoD2T seamlessly integrates traditional mathematical modeling with deep learning-based MOT frameworks, harnessing the advantages of both paradigms. A key innovation lies in the incorporation of a novel motion-static object tracking algorithm, grounded in motion distance and background features, thereby enhancing the precise determination of object motion status.

Furthermore, we propose the MVF1 metric to assess a model's ability to discern object motion states. We provide a concise analysis of MVF1, emphasizing its significance in comprehensively evaluating tracking performance.

This paper presents a fresh perspective in the field of multiple object tracking (MOT) by demonstrating the potential benefits of integrating traditional and deep learning methods. MoD2T not only addresses the limitations of various methods but also lays the groundwork for advancements in MOT within video analysis, owing to its remarkable capability in discerning object motion states. In future work, we intend to explore the integration of 3D object detection models. We believe that this approach could alleviate performance issues observed under rapid camera movement, particularly from the bird's-eye view (BEV) perspective.

	\bibliographystyle{IEEEtran}
	\bibliography{cangkao}
%	\bibliographystyle{IEEEtran}
%	\bibliography{IEEEexample}
	\newpage
	\newpage
	\newpage
	\section*{Appendices}
	{\appendices
		\section*{Traditional Method Double Gaussian Model}
		The process of traditional methods in this paper is very similar to  \cite{yiDetectionMovingObjects2013},which is a mixture of Gaussians, second version method(MOG2).
		
		Firstly, we use KIT\cite{heinrich2002inhibition} to correct the camera's motion.
		
		Then, we adopt a dual-mode single Gaussian model with three parameters: age, mean, and variance. The input image is divided into an $N*N$ grid, and each grid employs a single Gaussian model (SGM). This approach effectively reduces computational complexity while maintaining good detection performance.
		
		Mathematically, let $G_{i}^{(t)}$ represent the pixel group of grid $i$ at time $t$,$ \left| G_{i}^{(t)} \right|$ denotes the number of pixels in this group, and It represents the grayscale value of pixel $j$ at time $t$. The update functions for mean, variance, and age in the single Gaussian model are given by (17):
		
		\begin{equation}
			\begin{gathered}
				\mu_{i}^{(t)}=\frac{\tilde{\alpha}_{i}^{(t-1)}}{\tilde{\alpha}_{i}^{(t-1)}+1}\tilde{\mu}_{i}^{(t-1)}+\frac{1}{\tilde{\alpha}_{i}^{(t-1)}+1}M_{i}^{(t)}
				\\
				\sigma_{i}^{(t)}=\frac{\tilde{\alpha}_{i}^{(t-1)}}{\tilde{\alpha}_{i}^{(t-1)}+1}\tilde{\sigma}_{i}^{(t-1)}+\frac{1}{\tilde{\alpha}_{i}^{(t-1)}+1}V_{i}^{(t)}
				\\
				\alpha _{i}^{(t)}=\tilde{\alpha}_{i}^{(t-1)}+1
			\end{gathered}
		\end{equation}
		
		Here, $M_{i}^{(t)}$ denotes the average grayscale value within a grid at time $t$, and $V_{i}^{(t)}$ represents the squared difference between the moving average grayscale and the maximum grayscale value within a grid, as shown in (18) and (19):
		
		\begin{equation}
			M_{i}^{(t)}=\frac{1}{\left| G_i \right|}\sum_{j\in G_i}{I_{j}^{(t)}}
		\end{equation}
		\begin{equation}
			V_{i}^{(t)}=\max_{j\in G_i} \left( \mu _{i}^{(t)}-I_{j}^{(t)} \right) ^2
		\end{equation}
		
		If the variance exceeds a critical value (i.e.$, \tilde{\sigma}_{i}^{(t-1)}\,\,>\,\,\theta _v$), the model's age may need to be adjusted as shown in follow:
		\begin{equation}
			\tilde{\alpha}_{i}^{(t-1)}\gets \tilde{\alpha}_{i}^{(t-1)}\exp \left\{ -\lambda \left( \tilde{\sigma}_{i}^{(t-1)}-\theta _v \right) \right\} 
		\end{equation}
		
		We represent the parameters of the candidate model and the current model for grid $i$ at time $t$ as $\mu _{B,i}^{(t)},\sigma _{B,i}^{(t)} $and $\alpha _{B,i}^{(t)}$, respectively.If the observed difference between the average value $M_{i}^{(t)} $and the squared value $\mu _{B,i}^{(t)}$, is less than the critical value related to $\alpha _{B,i}^{(t)}$, as shown in 21:
		\begin{equation}
			\left( M_{i}^{(t)}-\mu _{A,i}^{(t)} \right) ^2<\theta _s\sigma _{A,i}^{(t)}
		\end{equation}
		
		where $\theta _s$ is a critical parameter. If the above condition is not satisfied but the observed average value matches the candidate background model, that is (22):
		
		\begin{equation}
			\left( M_{i}^{(t)}-\mu _{B,i}^{(t)} \right) ^2<\theta _s\sigma _{B,i}^{(t)}
		\end{equation}
		
		If neither of the two conditions mentioned above is met, the candidate background model is updated with the current observed value.
		
		At the same time, if (23) is satisfied:
		\begin{equation}
			\alpha _{B,i}^{(t)}>\alpha _{A,i}^{(t)}
		\end{equation}
		i.e., when the age of the candidate background model is greater than the current background model's age, the two models will exchange their roles.
		
		Finally, after determining the models, for the j-th pixel within the i-th pixel set, if  (24) is satisfied, the pixel is recognized as a foreground pixel, where d is a threshold value. Note that only the current background model is used to decide the foreground.
		
		\begin{equation}
			\left( I_{j}^{(t)}-\mu _{A,i}^{(t)} \right) ^2>\theta _d\sigma _{A,i}^{(t)}
		\end{equation}
		
		Once the foreground pixels are determined, applying some morphological operations yields bbox of the moving objects, as shown in the follow figure.
		
		\begin{figure}[htbp]
			\centering
			\includegraphics[width=3.5in]{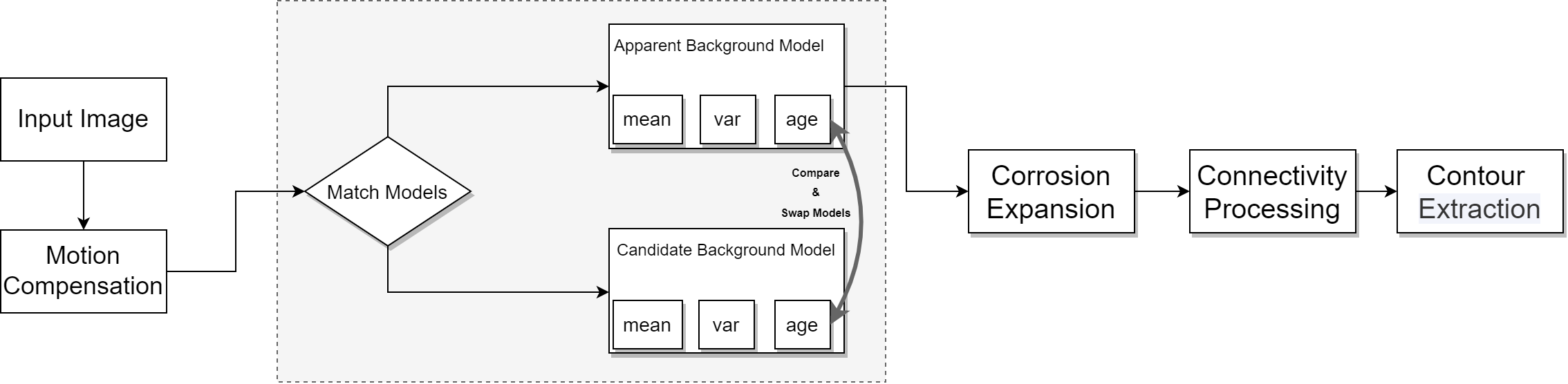}
			\caption{The framework of traditional method.}
			\label{tradition}
		\end{figure}

\end{document}